\documentclass[runningheads]{llncs}

\usepackage{appendix}
\usepackage[mobile]{eccv}

\renewcommand\paragraph{\@startsection{paragraph}{4}{\z@}%
                                    {0.5ex \@plus1ex \@minus.1ex}%
                                    {-1em}%
                                    {\normalfont\normalsize\bfseries}}

\usepackage{eccvabbrv}

\usepackage{graphicx}
\usepackage{booktabs}

\usepackage{multirow}
\usepackage{makecell}
\usepackage[accsupp]{axessibility}  %
\usepackage{wrapfig}

\usepackage{hyperref}

\usepackage{orcidlink}

\begin{document}

\title{Track Everything Everywhere Fast and Robustly } 

\titlerunning{CaDeX++}

\author{Yunzhou Song\inst{1}\thanks{Authors contributed equally to this work.} \and
Jiahui Lei\inst{1\star} \and
Ziyun Wang\inst{1} \and \\ Lingjie Liu\inst{1} \and Kostas Daniilidis\inst{1,2}
}

\authorrunning{Y. Song et al.}

\institute{University of Pennsylvania \and Archimedes, Athena RC\\
\email{\{timsong,leijh,ziyunw,lingjie.liu,kostas\}@cis.upenn.edu}}
\maketitle
\vspace{-2em}
\begin{center}
    \url{https://timsong412.github.io/FastOmniTrack/}
\end{center}
\vspace{-2em}
\begin{figure}
    \centering
    \includegraphics[width=1.0\linewidth]{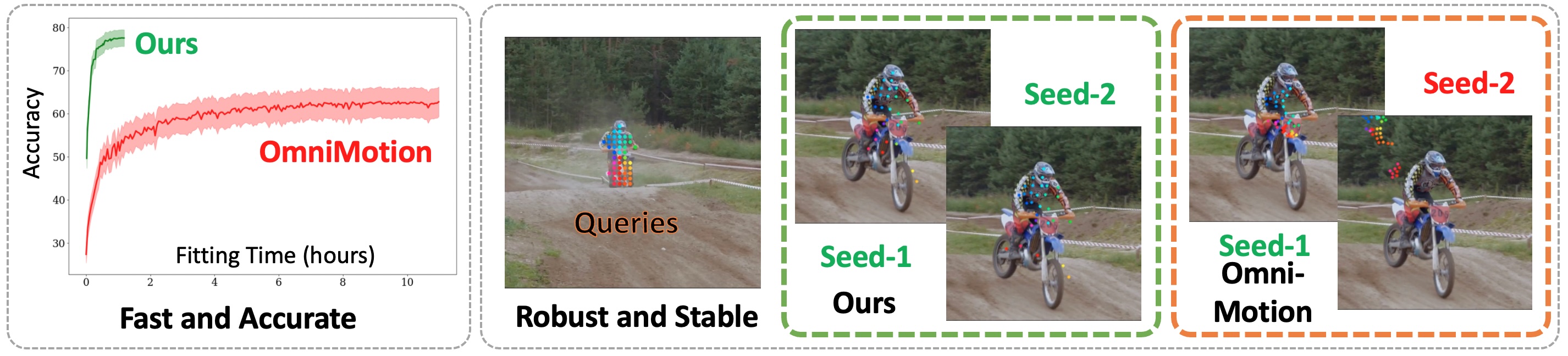}
    \caption{Our optimization-based approach achieves fast and robust long-term tracking}
    \label{fig:enter-label}
    \vspace{-3em}
\end{figure}

\begin{abstract}
We propose a novel test-time optimization approach for efficiently and robustly tracking any pixel at any time in a video. The latest state-of-the-art optimization-based tracking technique, OmniMotion~\cite{wang2023tracking}, requires a prohibitively long optimization time, rendering it impractical for downstream applications. OmniMotion~\cite{wang2023tracking} is sensitive to the choice of random seeds, leading to unstable convergence. To improve efficiency and robustness, we introduce a novel invertible deformation network, CaDeX++, which factorizes the function representation into a local spatial-temporal feature grid and enhances the expressivity of the coupling blocks with non-linear functions. While CaDeX++ incorporates a stronger geometric bias within its architectural design, it also takes advantage of the inductive bias provided by the vision foundation models. Our system utilizes monocular depth estimation to represent scene geometry and enhances the objective by incorporating DINOv2 long-term semantics to regulate the optimization process. Our experiments demonstrate a substantial improvement in training speed (more than \textbf{10 times} faster), robustness, and accuracy in tracking over the SoTA optimization-based method OmniMotion~\cite{wang2023tracking}.

\end{abstract}

\vspace{-3em}
\section{Introduction}
\vspace{-1em}
\label{sec:intro}

The association of visual information from continuous observations across long time horizons lays the foundation for modern spatial intelligence. In computer vision, one of the key tasks that provides this association is the long-term tracking of pixels,
which serves as the backbone for a wide spectrum of tasks, from 3D reconstruction to video recognition.

Previously, methods for estimating the correspondence can be divided into two categories based on their track representations. Feature-based methods represent points as local descriptors~\cite{rublee2011orb, lowe2004distinctive, bay2008speeded}, which can be matched over a long time horizon, due to the various invariance properties built into their design. However, feature descriptors are often sparsely matched due to the quadratic matching cost between every pair of images. On the other hand, optical flow methods estimate the motion of pixels in a dense manner~\cite{horn1981determining,  shi1994good,sun2018pwc,teed2020raft, xu2022gmflow, wang2019learning}. 
Due to the instantaneous nature of optical flow methods, they tend to perform poorly with long-range motion estimation and suffer from occlusion. Recently, several methods have been proposed to solve the problem via learning-based methods~\cite{karaev2023cotracker, harley2022particle, zheng2023pointodyssey, doersch2022tap, doersch2023tapir}. These methods learn strong prior knowledge by training on large synthetic datasets. In complement to learning-based methods, a new class of methods has emerged to optimize point tracks using test-time optimization on single scenes. A representative test-time optimization method is OmniMotion~\cite{wang2023tracking}, which is optimized to reconstruct a dynamic scene with a NeRF~\cite{mildenhall2020nerf,pumarola2021d} deformed by a global RealNVP~\cite{dinh2016density,lei2022cadex}, a normalizing flow network representing deformation. A major benefit of OmniMotion is that the optimization does not rely on strong prior knowledge, and is, thus, not susceptible to generalization gaps between training and testing.
However, due to the losses being only photometric, OmniMotion converges slowly when the training data do not provide enough constraints due to object and view occlusions. Moreover, the quality of reconstruction is often unpredictable because of the unconstrained random network initialization.

In this paper, we focus on advancing the computational efficiency, robustness, and accuracy of test-time optimization tracking methods~\cite{wang2023tracking} by introducing inductive bias through visual foundation models and network architecture.
One computational bottleneck of OmniMotion~\cite{wang2023tracking} is the cost of querying a global MLP-like NVP deformation network proposed first in CaDeX~\cite{lei2022cadex}. 
In Sec.~\ref{sec:method_local_nvp}, we introduce CaDeX++, a novel local feature-grid factorization of \textbf{invertible} deformation field, whose expressivity is further improved via a non-linear 1-D homeomorphism instead of the 1-D affine function in the NVP~\cite{lei2022cadex, wang2023tracking}. This design is inspired from NSVF~\cite{liu2020neural}, Instant-NGP~\cite{muller2022instant} and TensoRF~\cite{chen2022tensorf}, which exploit local factorized representations to boost global MLP-based NeRFs~\cite{mildenhall2020nerf}. 
Another time-consuming and under-constrained factor of OmniMotion~\cite{wang2023tracking} is the geometry reconstruction through volume rendering losses~\cite{mildenhall2020nerf}. Instead, (Sec.~\ref{sec:method_depth}) we regularize the optimization by initializing the optimizable per-frame depth map geometry based on monocular metric depth estimation, powered by the recent advances of 2D visual foundational models~\cite{bhat2023zoedepth}.
Finally, OmniMotion~\cite{wang2023tracking} only fits the short-term local optical flows~\cite{teed2020raft}, resulting in the lack of long-term association information. In Sec.~\ref{sec:method_dino}, we incorporate this missing information via incorporating the foundational DINOv2~\cite{oquab2023dinov2} feature correspondence into the fitting losses.
Leveraging a novel factorization of an invertible deformation field and vision foundation models as regularizers yields a novel method that achieves tracking accuracy and robustness improvement over OmniMotion~\cite{wang2023tracking}, while significantly reducing training time by more than \textbf{90\%}. 
We outline our contributions as follows.
\begin{itemize}
    \item An efficient and expressive novel invertible deformation network, CaDeX++, with local feature grid and non-linear interpolation.
    \item A novel depth-based geometry representation and the incorporation of DINOv2~\cite{oquab2023dinov2} long-term semantics, which boosts and regularizes the tracking optimization process.
    \item Significant speed up, stabilization, and performance improvement over OmniMotion~\cite{wang2023tracking} in long-term tracking task.
\end{itemize}

\vspace{-2em}
\section{Related Work}
\noindent\textbf{Pixel Tracking}:
Classical methods for estimating pixel correspondence can be divided into two categories: \textbf{keypoint tracking} and \textbf{optical flow}. For keypoint tracking methods, sparse feature descriptors are computed on local patches. Some common feature descriptors in visual odometry and SLAM methods include ORB~\cite{rublee2011orb}, SIFT~\cite{lowe2004distinctive}, and SURF~\cite{bay2008speeded}. Recently, a new class of methods has been proposed to learn feature descriptors using deep neural networks~\cite{zhang2014lift, detone2018superpoint}. The correspondence between two sets of feature descriptors can be matched using pairwise difference or using learned matching networks~\cite{li2023superglue}. Despite the different flavors of feature descriptors and matching algorithms, the correspondence is defined with respect to a predefined set of interest points. Detector-free methods~\cite{sun2021loftr} learn to match between all pairs of image locations without running feature detectors while having a global field of view. On the other hand, optical flow provides a dense correspondence field. In traditional optical flow computation, we often jointly optimize a data term and a regularization term. Horn and Schunck~\cite{horn1981determining} optimize a global flow field through gradient descent while adding a smoothness term to solve the classical aperture problem. Lucas and Kanade~\cite{lucas1981iterative} use the least squares criterion to optimize photoconsistency between flow-warped image patches and the new patches. This model solves an overdetermined system by assuming a parametric motion model. Later, learning-based approaches convert optimization terms into loss functions, allowing self-supervised optical flow training~\cite{sun2018pwc}. RAFT~\cite{teed2020raft} uses recurrent neural networks to simulate optimization steps in traditional optical flow. This architecture has been widely used to address optical problems due to its superior performance in handling different object and flow scales~\cite{xu2021high, jiang2021learning, zhang2021separable}. 

\noindent\textbf{Dense Long-range Tracking}
In the previous section, we provide an overview of two distinct types of tracking methods. Keypoint-based methods often provide correspondence over a longer time horizon, but the tracked points are usually sparse to save computational time. On the other hand, optical flow methods provide dense displacement of the pixels but fall short in long-term tracking consistency. Particle Video~\cite{sand2008particle} is proposed to optimize long-range motion while preserving the density of optical flow estimation, by connecting short-term flow and regularizing the distortion between particles. Recently, PIPs~\cite{harley2022particle} built on the original particle videos by proposing a deep MLP-Mixer module to iteratively update the long-term tracks. TAP-Net~\cite{doersch2022tap} uses a small neural network to directly regress point locations. TAPIR~\cite{doersch2023tapir} starts with the TAP-Net initialization and refines the point trajectories with the MLP-Mixer architecture of PIPs. MFT~\cite{neoral2024mft} estimates the flow uncertainty and occlusion maps, which are used to select high-confidence flow chains to generate long-term tracks. In PointOdyssey~\cite{zheng2023pointodyssey}, Zheng et al. design PIPs$++$ to increase the temporal field of view using a convolution over time and memorize the most recent appearance templates.
OmniMotion~\cite{wang2023tracking} optimizes the per-pixel point tracks by lifting the 2D pixels into 3D and fitting optimizeable invertible warping functions. This representation allows for flexible tracking of long videos. The modeled scene is represented as a canonical 3D volume with its bijective mappings to each quasi-3D local scene. Both the canonical space and the mapping functions are jointly optimized.

\vspace{-1em}
\section{Method}

\begin{figure}[t]
    \centering
    \includegraphics[width=0.8\linewidth]{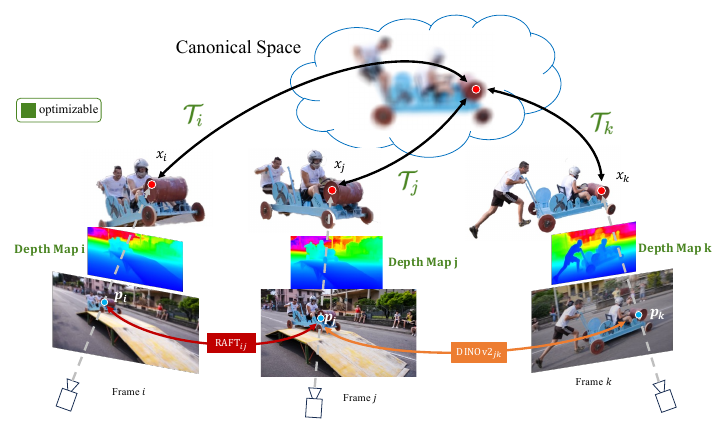}
    \caption{Method Overview: To track a query pixel $p_i$, we first lift the pixel to 3D with an optimizable depth map (Sec.~\ref{sec:method_depth}). The 3D point is deformed into the shared canonical space and back to another time frame $j$ with a novel efficient and expressive invertible deformation field $\mathcal T$ (Sec.~\ref{sec:method_local_nvp}). 
    The depth maps and the deformation $\mathcal T$ are optimized with both short-term dense RAFT~\cite{teed2020raft} optical flow and long-term sparse DINOv2~\cite{oquab2023dinov2} correspondence (Sec.~\ref{sec:method_dino}).}
    \label{fig:main}
\end{figure}

\vspace{-0.5em}
\subsection{Preliminaries}
\label{sec:method_pre}
Given an RGB video sequence $\{I_t\}_{t=1}^T$ with $T$ frames and an arbitrary query pixel $ p_t \in \mathbb{R}^2$ from a video frame $I_t$, our goal is to predict its long-term sequential trajectories $(\hat{p}_1, ..., \hat{p}_T)$ as well as the visibility $(\hat{v}_1, ..., 
\hat{v}_T) \in \{0, 1\} ^T$. 
In the following sections, we will use subscripts to indicate the time or frame, and superscripts to indicate the track identity. An overview of our method is in Fig.~\ref{fig:main}.

A key consideration for an optimization-based long-term tracker is designing the parameterization of long-term tracks $(\hat{p}_j, \hat{v}_j) = \mathcal F(  {p_i}, j)$, where $\mathcal F$ is the model of long-term tracks that takes input any query pixel $  {p_i}$ and destination time $j$, predicts the position $\hat{  {p}}_j$ and visibility $\hat{v}_j$ at time $j$.
The current SoTA OmniMotion~\cite{wang2023tracking} parameterizes this tracking function $\mathcal F$ as a rendering process~\cite{wang2023tracking}. First, the query pixel $p_i$ at time $i$ is marched along a line of points on the ray $x_i^k = o(p_i) + z^k d(p_i)$ where $o$, $d$ and $z$ are the ray center, direction, and marching depth, respectively. An invertible deformation field~\cite{lei2022cadex} $u = \mathcal T_i(x_i)$ is used to deform each ray-marching position at time $i$ to the shared global canonical space position $u$, where the geometry and appearance of the scene are modeled as a canonical radiance field $(\text{color}, \sigma)=G(u)$. Finally, all the canonical positions on the bent ray are mapped back to the target time frame $j$ with the inverse of the deformation field $\mathcal T_j^{-1}$:
\begin{equation}
    x_j^k = \mathcal T_j^{-1}(u^k) = \mathcal T_j^{-1} \circ \mathcal T_i (x_i^k).
    \label{eq:omni_corr}
\end{equation}
The prediction of the target pixel position can be formulated as a rendering process:
\begin{equation}
    \hat p_j = \pi\left(\sum_{k=1}^K T_k \alpha_k x_j^k\right), \quad T_k=\prod_{l=1}^{k-1}(1-\alpha_l), \quad \alpha_k = 1-\exp(1-\sigma_k),
    \label{eq:omni_rendering}
\end{equation}
where $\pi$ is the camera projection function and $\sigma$ is the opacity predicted by the canonical radiance field $G$.
The noisy short-term optical flow pairs $\mathcal P_{\text{RAFT}} = \{(p_i, p_j) \; i, j \in [1, ..., T]\}$ are usually assumed given as optimization targets, which are predicted by well-established networks like RAFT~\cite{teed2020raft}. OmniMotion composes the tracking function above with a set of learnable $G,\mathcal T$ networks, and fits it against the noisy local optical flow pairs $\mathcal P$, while minimizing the rendering photometric errors.
Similarly, the visibility can be found in the rendering process. For further information, readers are directed to Wang et al.~\cite{wang2023tracking}. We will see in the next sections why Eqs.~\ref{eq:omni_corr},\ref{eq:omni_rendering} are inefficient and result in high-variant fittings and how we address these issues.

\subsection{CaDeX++: Non-linear and Local Invertible NVPs}
\label{sec:method_local_nvp}

\begin{figure}[t]
    \centering
    \includegraphics[width=1.0\linewidth]{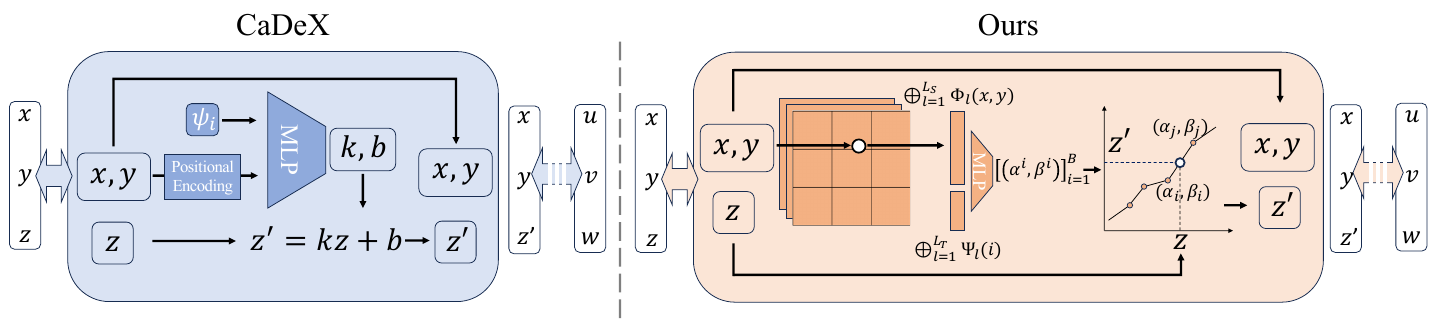}
    \caption{Architecture of CaDeX++ (right). The deformation network has a stack of coupling blocks and gradually changes one coordinate dimension per block (For difference Sec.~\ref{sec:method_local_nvp}).
    }
    \label{fig:cadex++}
    \vspace{-2em}
\end{figure}

The expressivity and efficiency of $\mathcal T$ in Eq.~\ref{eq:omni_corr} are critical when we extensively query the deformation field to model the long-track function $\mathcal F$. As shown in Fig.~\ref{fig:cadex++}-Left, OmniMotion\cite{wang2023tracking} uses CaDeX~\cite{lei2022cadex}, a \textbf{global} NVP to parameterize $\mathcal T$, which consists of a stack of coupling blocks. 
During each coupling iteration, a single dimension of the coordinates, such as $z$, is modified by a global MLP queried by the other two coordinates ($x$ and $y$ when $z$ is modified) and a global latent code.
To ensure the invertibility of this coupling step, $z$ is changed by a simple 1-D affine mapping predicted by the MLP in the following coupling block:
\begin{equation}
    (k, b) = \text{MLP}\left([x,y];   {\psi_i}\right), \quad z' = kz+b
    \label{eq:cadex_nvp}
\end{equation}
where $\psi_i$ is a time-dependent global latent code. A stack of such coupling blocks parameterized by per-block MLPs alternatingly changes the coordinates gradually. Please see CaDeX~\cite{lei2022cadex} and OmniMotion~\cite{wang2023tracking} for details.

However, the NVP~\cite{lei2022cadex} formulation in Eq.~\ref{eq:cadex_nvp} has two main drawbacks, which we overcome with CaDeX$++$.
First, the MLP and the latent codes are all global, which requires large networks for sufficient capacity.
Inspired by global MLP-based NeRF versus local feature-grid-based representations, such as Instant-NGP~\cite{muller2022instant} and TensoRF~\cite{chen2022tensorf}, we ask the question: Can we factorize the \textbf{invertible} deformation field~\cite{lei2022cadex} into \textbf{local} representations as well? At first sight, achieving invertibility may seem challenging due to the need for a specific network structure. 
We propose a novel approach to exploiting the desired locality by factorizing the latent code $\psi$ while significantly reducing the $MLP$ network size. Specifically, the latent code $\psi$ that controls the coordinate deformation of each coupling block can be factorized into a multi-resolution lookup function. For example, we can factorize $\psi$ in Eq.~\ref{eq:cadex_nvp}, indexed by unchanged coordinates, $x,y$ and the time index $i$ as
\begin{equation}
    \psi(x,y,i) = \left(\oplus_{l=1}^{L_T}{\Psi_l(i)}\right) \oplus \left(\oplus_{l=1}^{L_S}{\Phi_l(x,y)}\right),
    \label{eq:ours_local}
\end{equation}
where $\oplus$ denotes feature concatenation and $L_T,L_S$ are the spatial and temporal feature grid resolution levels, respectively. 
$\Psi_l(i)$ and $\Phi_l(x,y)$ are bi-linearly querying a 1-D or 2-D feature grid at resolution $l$, respectively. Eq.~\ref{eq:ours_local} is a local spatial-temporal factorization that decouples time and space. Note that when we replace $\phi_i$ in Eq.~\ref{eq:cadex_nvp} with $\psi(x,y,i)$ from Eq.~\ref{eq:ours_local}, the invertibility still holds since $\psi(x,y,i)$ does not depend on the changing $z$ coordinate.

Another drawback of CaDeX~\cite{lei2022cadex} in Eq.~\ref{eq:cadex_nvp} is the insufficient expressivity of the affine function applied to the changing $z$ dimension.
The only requirement for invertibility is to ensure that the function that changes $z$ is invertible and the affine function of the form $kz+b$ is the simplest among all such functions.
To increase the expressivity within the limited number of coupling blocks, we propose to use the monotonic piece-wise functions as non-linear deformation. 
Specifically, the 1D function is parameterized by a list of $B$ control points $[(\alpha^1,\beta^1), \ldots (\alpha^B,\beta^B)]$ with piece-wise linear interpolation:
\begin{equation}
    z' = \frac{z-\alpha^i}{\alpha^j-\alpha^i} (\beta^j-\beta^i)+\beta^i, \; z\in[\alpha^i, \alpha^j),\;  j - i = 1.
\label{eq:ours_nonlinear}
\end{equation}
To guarantee the monotonicity of the control points, we make the network predict the positive delta values as:
\begin{equation}
    [(\Delta \alpha^1, \Delta \beta^1)\ldots, (\Delta \alpha^{B}, \Delta \beta^{B})] = \text{TinyMLP}\left([x,y]; \psi(x,y,i)\right),
    \label{eq:ours_deformation_summary}
\end{equation}
where $\Delta \alpha^b>0$ and $\Delta \beta^b>0$. For further information regarding the interpolation and network structures, please refer to our supplementary document.
By incorporating locality and non-linearity inductive bias, we enhance both efficiency and expressiveness, all while preserving the essential guarantees of invertible characteristics.
\vspace{-1.5em}
\subsection{Optimization with Depth Prior}
\vspace{-0.5em}
\label{sec:method_depth}
Although we model the deformation field efficiently with CaDeX$++$, the optimization process of OmniMotion~\cite{wang2023tracking} can often be unstable and slow. The undesirable optimization performance arises from the scene's geometry being optimized using a volume rendering loss as described in Eq.~\ref{eq:omni_rendering}.
Moreover, with the small camera baseline in many casual videos, a standard NeRF~\cite{mildenhall2020nerf}- may lead to a reconstruction that is highly ambiguous because the accuracy of the "triangulation" of photometric loss is compromised by the limited parallax.
Therefore, we avoid such a NeRF-like reconstruction process by explicitly exploiting recent advances in foundational monocular metric depth estimation, i.e. ZoeDepth~\cite{bhat2023zoedepth}, which estimates a reasonably accurate and consistent geometry for each frame. Note that we use the \textbf{metric} depth models~\cite{bhat2023zoedepth, depthanything, Guizilini_2023_ICCV} as opposed to a scale-invariant depth models~\cite{ke2023repurposing,birkl2023midas, Xian_2018_CVPR} to avoid inconsistency of scale within a video.

Given an initial depth map $D_i$ estimated from ZoeDepth for every video frame, the tracking function $\mathcal F$ in Eq.~\ref{eq:omni_corr} simply reduces to back-projection, deformation, and projection:
\begin{equation}
    \hat p_j = \pi\left(
        \mathcal T_j^{-1} \circ \mathcal T_i ( \pi^{-1}(D_i[p_i], p_i))
        \label{eq:ours_corr}
    \right),
\end{equation}
where $\pi^{-1}$ is the back-projection function that lifts the query pixel $p_i$ with its depth $D_i[p_i]$ into 3D. Note that the projection distortion can be effectively absorbed into $\mathcal T$ because the deformation is learnable.
We follow OmniMotion~\cite{wang2023tracking} to use a fixed pin-hole camera with a FOV of $40$ degrees. 
Given the inaccuracy of the depth maps $D_i$ obtained from ZoeDepth, we set all $D_i$ \textbf{optimizable}, regularized by a smoothness term, as detailed in Section~\ref{sec:method_inference}. 
In summary, the inefficient and under-constrained radiance field $G$ in Eq.~\ref{eq:omni_rendering} is replaced with a list of optimizable depth maps $\{D_i\}_{i=1}^T$ to boost and stabilize the optimization process.

\vspace{-1.5em}
\subsection{Incorporation of Long-term Semantics}
\label{sec:method_dino}
\setlength{\intextsep}{1pt}
\begin{wrapfigure}{r}{0.5\textwidth}
  \centering
    \includegraphics[width=0.5\textwidth]{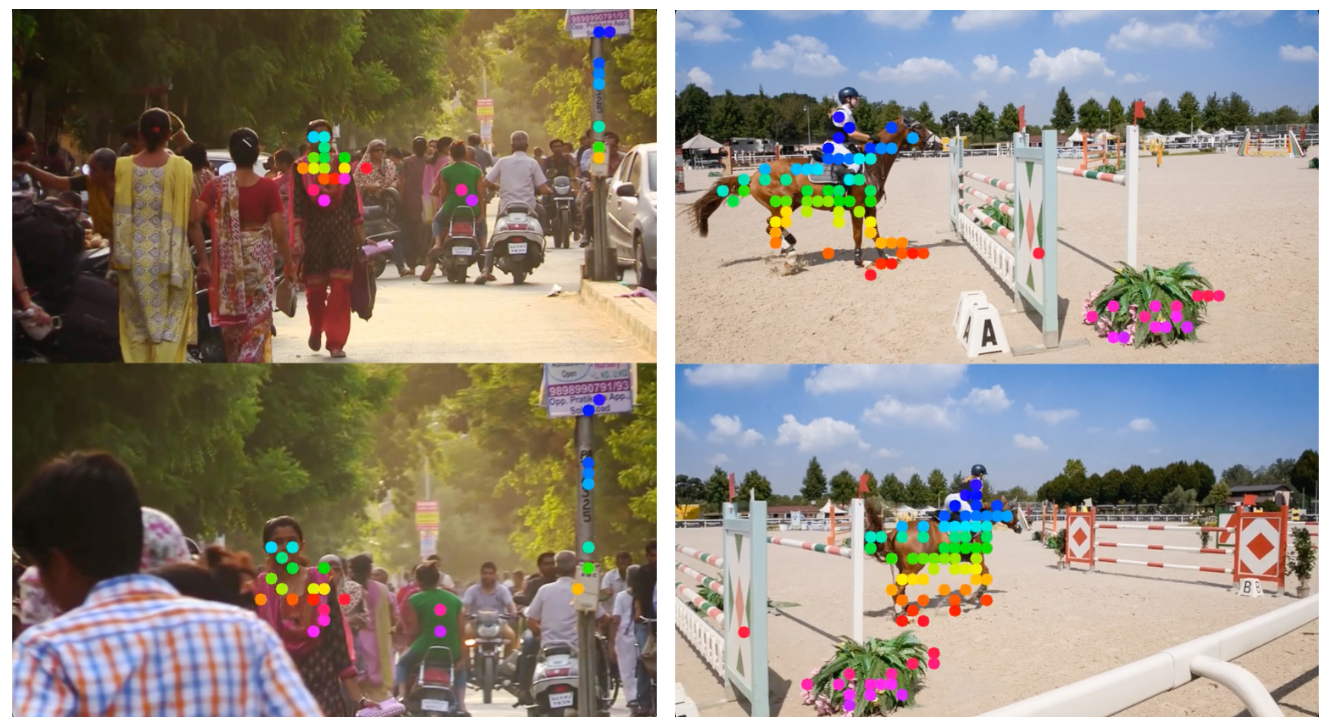}
    \caption{Filtered long-range semantic correspondences based on DINOv2~\cite{oquab2023dinov2}.}
    \label{fig:dino_pairs}
\end{wrapfigure}
To optimize $\mathcal T$, OmniMotion~\cite{wang2023tracking} relies solely on short-term optical flow as the fitting target. Inspired by recent progress in 2D visual foundational features, we incorporate sparse long-term semantic correspondence into the optimization targets, by using image features pre-trained on large image datasets.
Specifically, we utilize and filter the DINOv2~\cite{oquab2023dinov2} features to establish long-term correspondences that are sparse but reliable.
Given two DINOv2 feature maps $F_i, F_j$, we first compute the inter-frame pairwise cosine similarity for every two patch features between $i,j$ and choose the mutually consistent nearest neighbor matches as candidates for long-term correspondence. This cycle consistency criterion means that a patch $i$ whose best match is patch $j$ must also be the closest match of patch $j$.
We then filter the candidates by evaluating the self-similarity within each frame of the feature map. This helps us avoid any ambiguity in matching due to the absence of texture and noise in the feature map. Please refer to the additional materials for further information on the filtering process.
In summary, through the utilization of DINOv2~\cite{oquab2023dinov2}, we augment the initial optical flow optimization objectives in OmniMotion~\cite{wang2023tracking} by incorporating a broader range that encompasses long-term sparse correspondence.

\vspace{-1em}
\subsection{Training and Inference}
\label{sec:method_inference}
During training (test-time optimization), given a pair of 2D correspondence from the target sets $(p_i, p_j) \in \mathcal P = \mathcal P_{\text{RAFT}} \bigcup  \mathcal P_{\text{DINOv2}}$,  we randomly choose one pixel as the query and another as the target. 
For the query pixel $p_i$, we back-project the $p_i$ into 3D by looking up its depth from the optimizable depth map $D_i$ as $x_i = \pi^{-1}(D_i[p_i], p_i)$. 
We then map $x_i$ directly to time $j$ by $\hat{x}_{i\rightarrow j} = \mathcal{T}^{-1}_j(\mathcal{T}_i(x_i))$ and project it to 2D screen to get the prediction pixel coordinate $\hat{p}_{i\rightarrow j} = \pi(\hat{x}_{i\rightarrow j})$ as in Eq.~\ref{eq:ours_corr}. 
We define the losses between $  {\hat{{p}}}_{i\rightarrow j}$ and $p_j$ as follows:
\begin{itemize}
    \item  \textbf{Pixel Position Loss}: We minimize the mean absolute error for both flow supervision points and long-term matching supervision points denoted as $\mathcal{L}_{p}$: 
    \begin{equation}
        \mathcal{L}_{p} =\frac{1}{|\mathcal P|} \sum_{(p_i, p_j) \in \mathcal P}||\hat{p}_{i\rightarrow j} - p_j||_1
        \label{eq:loss_pixel}
    \end{equation}
    where  $\mathcal P = \mathcal P_{\text{RAFT}} \bigcup  \mathcal P_{\text{DINOv2}}$ is the set of all correspondence generated by optical flow and long-term semantics.

    \item \textbf{Depth Consistency Loss}: Since the depth maps initialized from ZoeDepth~\cite{bhat2023zoedepth} are not perfectly accurate, we additionally supervise the deformed point $\hat{x}_{i\rightarrow j}$ depth consistency with the target pixel's optimizable depth $D_j[p_j]$:
    \begin{equation}
        \mathcal{L}_{d} = \frac{1}{|\mathcal P|} \sum_{(p_i, p_j) \in \mathcal P} ||z(  {\hat x}_{i\rightarrow j}) - D_j[p_j]||_1
        \label{eq:loss_depth}
    \end{equation}
    \item \textbf{Depth Regularization Loss}: To ensure stability for depth optimization, we restrict the depth maps that can be optimized to remain near the initially set depth map. Given the initial ZoeDepth~\cite{bhat2023zoedepth} depth map predictions $D_i^{\text{init}}$ and their spatial gradients $\nabla_{p}D_i^{\text{init}}$, we regularize the optimized depth maps to stay close to the initialization:
    \begin{equation}
    \mathcal{L}_{\text{reg}} = \frac{1}{|\mathcal P|} \sum_{(p_i, p_j) \in \mathcal P}  ||\nabla_{p}D_j^{\text{init}}[p_j] - \nabla_{p}D_j[p_j]||_2 + ||D_j^{\text{init}}[p_j] -D_j[p_j]||_1 
    \label{eq:loss_reg}
    \end{equation}
\end{itemize}

\noindent The final total loss is the weighted sum of the loss terms above:
\begin{equation}
    \mathcal{L} = \mathcal{L}_{p} + \lambda_d\mathcal{L}_{d} + \lambda_{\text{reg}} \mathcal L_{\text{reg}},
\end{equation}
where $\lambda_p, \lambda_d, \lambda_{reg}$ are the loss balancing weights.

During inference, the long-term trajectory is efficiently predicted by Eq.~\ref{eq:ours_corr} given any query position.
For the visibility $\hat v_j$ at target time $j$, we simply compare the $z$ value of the warped 3D point $\hat x_{i\rightarrow j}$ from query time $i$ with $D_j[\hat p_{i\rightarrow j}]$, the depth value at frame $j$. Occlusion is detected if the warped 3D point is behind the depth value than a small threshold $\epsilon_d$.

\vspace{-1.5em}
\section{Experiments}
\vspace{-0.5em}

\subsection{Experiment Setup}
\vspace{-0.5em}
\label{sec:exp_setup}
\noindent\textbf{Dataset}: Following OmniMotion~\cite{wang2023tracking}, we evaluated our method on the following datasets from TAP-Vid~\cite{doersch2022tap}: 
\vspace{-0.7em}
\begin{itemize}
    \item \textbf{DAVIS}~\cite{davis}, a real scene dataset of 30 videos from the DAVIS 2017 validation set. Each video contains 34 to 104 RGB frames. In this dataset, we observe both camera and scene motions.
    \item \textbf{RGB-Stacking}~\cite{rgbstacking}, a synthetic robot manipulation dataset. The dataset is composed of 50 videos, each with 250 RGB frames. The videos are rendered with only object motion with a static camera.
\end{itemize}
\vspace{-0.7em}
\noindent\textbf{Metrics}:
\vspace{-0.7em}
\begin{itemize}
    \item \textbf{$\delta^x_\mathbf{avg}$} The average position precision percentage of tracked points that fall within $x$ absolute pixel error of their targets. The metric is defined for all points that are visible in the ground truth. It has 5 thresholds $\delta^x, \; x \in \{1, 2, 4, 8, 16\} $, where $\delta^x$ is the fractions of points that lie within $x$ pixels of their ground truth position.
    \item \textbf{Average Jaccard (AJ)} The joint accuracy of points that are ground-truth visible. It measures the mean proportion of points that both lie within $x$ pixels of their ground truth position and are predicted as visible.
    \item \textbf{Occlusion Accuracy (OA)} The fraction of the correct visibility prediction for all points in a frame. The numerator is the number of correct predictions including both visible and occluded points.
    \item \textbf{Temporal Coherence (TC)} The mean $L_2$ distance between the acceleration of actual tracks and predicted tracks is determined by calculating the difference in flow between three consecutive frames $i, j, k$ for visible points, denoted as $f_{j \rightarrow k} - f_{i\rightarrow j}$.
\end{itemize}

We conducted all experiments on 480p images and evaluated metrics on 256x256 images following the training and evaluation protocols of OmniMotion~\cite{wang2023tracking}.

\begin{figure}[h]
    \centering
    \includegraphics[width=0.8\linewidth]{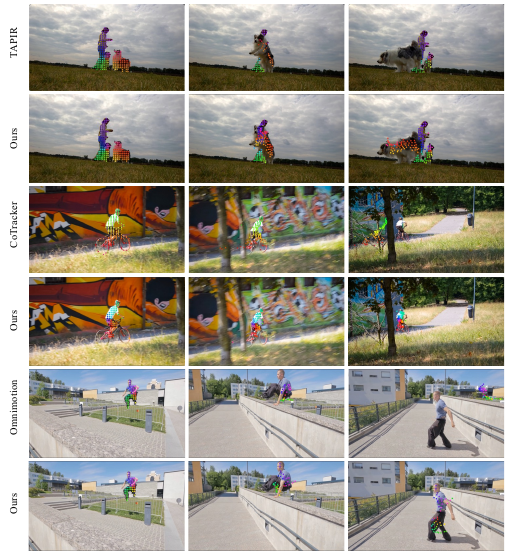}
    \caption{We compare the tracking performance our method with TAPIR~\cite{doersch2023tapir}, Cotracker~\cite{karaev2023cotracker} and OmniMotion~\cite{wang2023tracking} on DAVIS scenes \textit{dogs-jump, bmx-trees}, and \textit{parkour} from top to bottom. The leftmost column shows the initial query points. Our method performs better on these scenes than the other method. }
    \label{fig:main-qual}
    \vspace{-2em}
\end{figure}

\vspace{-1.5em}
\subsection{Comparison with SoTA Methods}
\vspace{-0.5em}
\label{sec:exp_comp}

\subsubsection{Baselines}We compare our method with feed-forward methods and optimization-based methods. Some of the representative baselines are: 
\textbf{1) PIPs}~\cite{harley2022particle} is a method that iteratively updates the position and visibility of a trajectory point within 8 frames. The long-term trajectories are obtained by zipping overlapping windows.
\textbf{2) TAP-Net}~\cite{doersch2022tap} is a simple baseline that computes correspondence by directly querying the feature cost volume of a pretrained visual backbone. 
\textbf{3) TAPIR}~\cite{doersch2023tapir} initializes a trajectory using an exhaustive global matching process and refines the point location, occlusion, and uncertainty iteratively with local features.  
\textbf{4) CoTracker}~\cite{karaev2023cotracker} is the state-of-the-art long-term tracking method. Cotracker updates several trajectories jointly by computing cross-track/time attention, allowing trajectory prediction with a global receptive field over all tracked points.      
\textbf{5) OmniMotion}~\cite{wang2023tracking} is a test-time optimization method that optimizes point correspondences using a set of invertible mapping functions between each frame to canonical space. The underlying representation is an optimizable NeRF volume. 

\vspace{-1.5em}
\subsubsection{Quantitative comparisons} We present the quantitative evaluation results in Tab. \ref{tab:headings}. Our method achieves the best temporal coherence among all methods on DAVIS and has better position precision than other optimizable methods, which is comparable with other feed-forward methods. On RGB-stacking our method performs better than other feed-forward methods. 

Compared to pure flow-based optimization approaches, our method achieves significantly better precision and temporal coherence on complex motions over the real scene dataset. Our method incorporates long-term supervision with short-term ones, which simultaneously corrects the global trajectory coarsely and refines the detailed motion locally. Compared with feed-forward approaches, our method achieves better on the textureless synthetic videos. Feature-based methods rely on visual textures to track contrastive points, which are prone to fail when tracking multiple identical points. More analyses are specified in section \ref{sec: further}.

\begin{figure}[tb]
    \centering
    \includegraphics[width=1\linewidth]{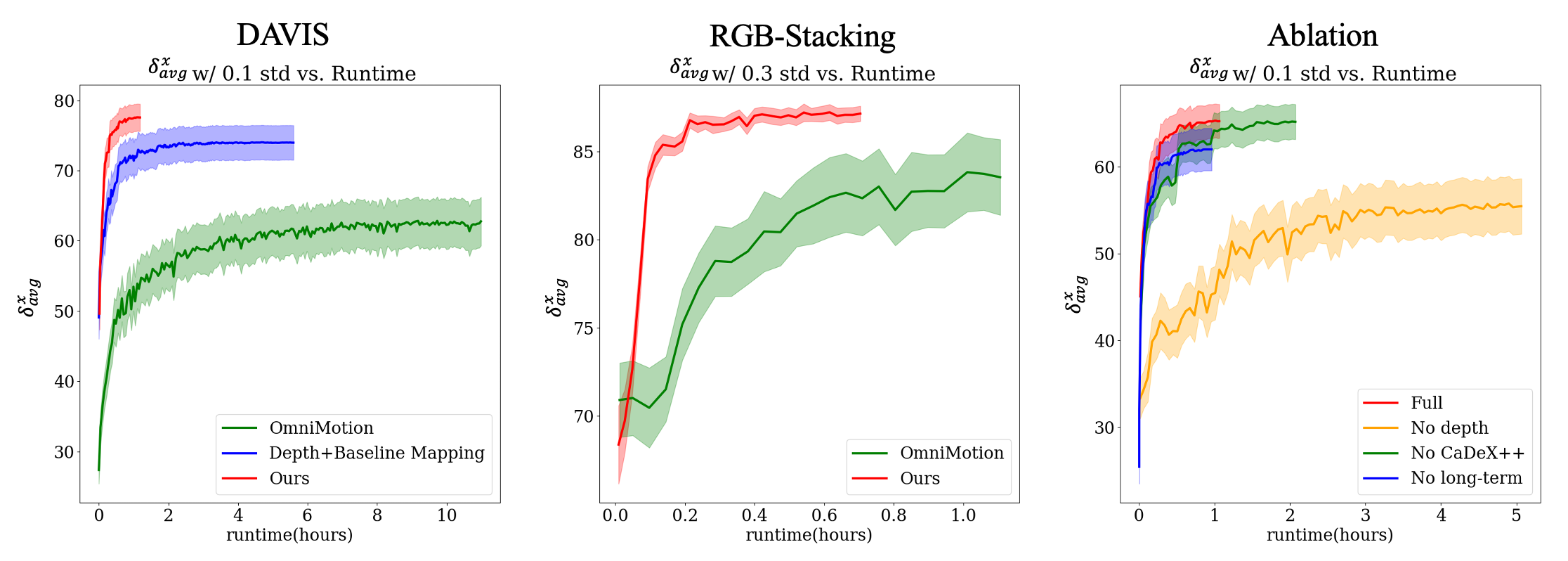}
    \caption{Runtime Comparisons for DAVIS, subset of RGB-Stacking, and ablation experiments}
    \label{fig: runtime}
\end{figure}

Without neural rendering for depth or color and equipped with the novel CaDeX++, our method accelerates the convergence more than 10 times faster than OmniMotion on DAVIS and 5 times faster on RGB-stacking approximately as shown in Fig. \ref{fig: runtime}. We conduct the experiments on NVIDIA V100 GPUs. 
\vspace{-1em}
\subsubsection{Qualitative comparison} Fig. \ref{fig:main-qual} reveals that compared with baselines, our method can track points against long-term occlusion. Our method can also handle complex object motion and large camera motion.

\begin{table}[tb]

  \setlength{\tabcolsep}{3pt}
  \caption{Quantitative comparison of our method and baselines. We categorized all methods into two categories: the feedforward methods which first train a network and then inference trajectories on testing videos, and the optimization-based methods which fuse pairs of pixel correspondence into trajectories for each testing scene without a pre-trained tracking network. 
  }
  \label{tab:headings}
  \centering
  \begin{tabular}{llccccccccc}
    \toprule
    \multicolumn{2}{c}{\textbf{Method}} & \multicolumn{4}{c}{\textbf{DAVIS}} & \multicolumn{4}{c}{\textbf{RGB-Stacking}}\\
     \cmidrule(r){3-6}   \cmidrule(r){7-10} 
     &  & AJ$\uparrow$ & $\delta^x_{avg}$ $\uparrow$ & OA$\uparrow$  & TC $\downarrow$ & AJ $\uparrow$ & $\delta^x_{avg}$ $\uparrow$ & OA$\uparrow$ & TC $\downarrow$\\

    \midrule
    
    \multirow{6}{*}{\makecell[c]{Feed-\\forward}} & PIPs~\cite{harley2022particle}    & 39.9  &   56.0    & 81.3  & 1.78 &   37.3  &  50.6 &  89.7 & 0.84\\

    &   Flow-Walk~\cite{bian2022learning} & 35.2 & 51.4 & 80.6 & \textbf{0.90} & 41.3 & 55.7 & 92.2 & \textbf{0.13} \\

     &   MFT~\cite{neoral2024mft} &   56.1    &   70.8    &   86.9    &   -   &   -   &   -   &   -   &   -    \\
    &   TAP-Net~\cite{doersch2022tap} & 38.4 & 53.4 & 81.4 & 10.82 & 61.3 & 73.7 & 91.5 & 1.52 \\
    
    &   TAPIR~\cite{doersch2023tapir}   &   59.8 & 72.3 & 87.6 &  -   &   \textbf{66.2}    &   77.4    & \textbf{93.3}  & - \\
    
    &   CoTracker~\cite{karaev2023cotracker}   &   \textbf{65.1}    &   \textbf{79.0}    &   \textbf{89.4}    &   0.93 & 65.9 &    \textbf{80.4} &  85.4    &   0.14 \\
    
    \midrule
    
    \multirow{4}{*}{\makecell[c]{Opti-\\mization}}   &   Connect RAFT~\cite{teed2020raft} & 30.7 & 46.6 & 80.2 & 0.93& 42.0 & 56.4 & 91.5  & 0.18\\
    
    & Deformable Sprites~\cite{ye2022deformable} & 20.6 & 32.9 & 69.7 & 2.07 & 45.0 & 58.3 & 84.0 & 0.99 \\
   
    &   OmniMotion~\cite{wang2023tracking} &   51.7    &   67.5    & 85.3  & 0.74 &  \textbf{77.5}  &  87.0 &  93.5 & \textbf{0.13} \\
    &   Ours    &  \textbf{59.4}    &   \textbf{77.4 }&     \textbf{85.9} & \textbf{0.68} & 75.4 & \textbf{87.1} & \textbf{93.6} & 0.15 \\

  \bottomrule
  
  \end{tabular}
\end{table}

\vspace{-1.5em}
\subsection{Ablation Study}
\vspace{-0.5em}
\begin{figure}[tb]
    \centering
    \includegraphics[width=1.0\linewidth]{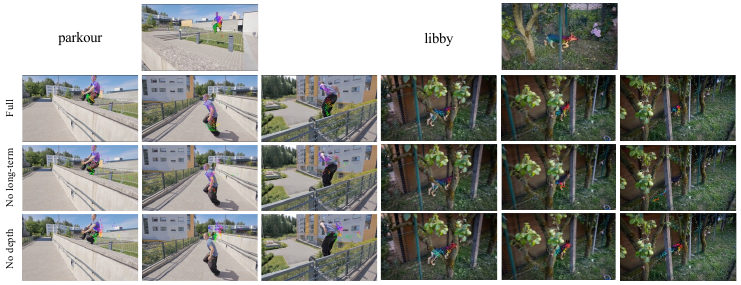}
    \caption{Qualitative comparison of ablation configurations. }
    \label{fig: qual_abla}
    \vspace{-1em}
\end{figure}

We perform ablations to verify our design decisions listed in Tab. \ref{tab: ablation} on a subset of the DAVIS~\cite{davis} dataset. \textit{No depth} indicates replacing the optimizable depth maps with photometric neural rendering to predict depth. \textit{No long-term} is a version that excludes long-term supervision in the training dataset. \textit{No CaDeX++} is the model that downgrades the local invertible mapping into the baseline global MLP ones.

As shown in Tab. \ref{tab: ablation} and Fig. \ref{fig: runtime}, the introduction of the optimizable depth maps significantly improves the tracking precision and converging speed by leveraging ordinal information from the depth priors to cluster depth semantically. Long-term supervision enhances the trajectory precision considerably and CaDeX$++$ accelerates convergence speed.

Qualitative results demonstrated in Fig. \ref{fig: qual_abla} prove that the introduction of the depth prior makes the tracking of points within the same instance more concentrated and less prone to dispersion. Besides, without long-term supervision, our method fails to handle large and frequent occlusions across time.

\begin{table}[tb]
    \begin{minipage}{.47\linewidth}
        \caption{Ablation study on a subset of DAVIS. We ablate loss two loss terms and CaDex++ architecture.}
        \label{tab: ablation}
      \centering
      \begin{tabular}{lcccc}
        \toprule
        \textbf{Method} & AJ~$\uparrow$ & $\delta^x_{avg}$~$\uparrow$ & OA~$\uparrow$ & TC~$\downarrow$  \\ 
        \midrule
        No depth & 42.0 & 56.8 & 73.3 & 1.42 \\
        No long-term & 45.6 & 61.3 & 75.5 & 1.32\\
        No CaDeX++ & 48.2 & 65.4 & 80.1 &  $\mathbf{0.97}$\\
        \midrule
        Full & $\mathbf{48.6}$ & $\mathbf{65.7}$ & $\mathbf{80.1}$ & 1.14 \\
        \bottomrule
        \end{tabular}
    \end{minipage}%
    ~
    \begin{minipage}{.47\linewidth}
      \centering
        \caption{Disagreement between tracking trajectory and optical flow. Lower is better, indicating better consistency.}
        \begin{tabular}{p{1.5cm}ccc}
        \toprule
        
        \textbf{Method} &  \multicolumn{2}{c}{\textbf{DAG}$\downarrow$} \\
                    \cmidrule(r){2-3}
                    &    car-turn & plane \\ 
        \midrule
        CoTracker    & 40.3 &  32.5 \\ 
        Ours   &   \textbf{14.9}   &   \textbf{12.8} \\
        \bottomrule
        \end{tabular}
        \label{tab: agreement}
    \end{minipage} 
\end{table}

\vspace{-1.5em}
\subsection{Further Comparison with CoTracker and OmniMotion}
\vspace{-1em}
\label{sec: further}

\begin{figure}[tb]
    \centering
    \includegraphics[width=1.0\linewidth]{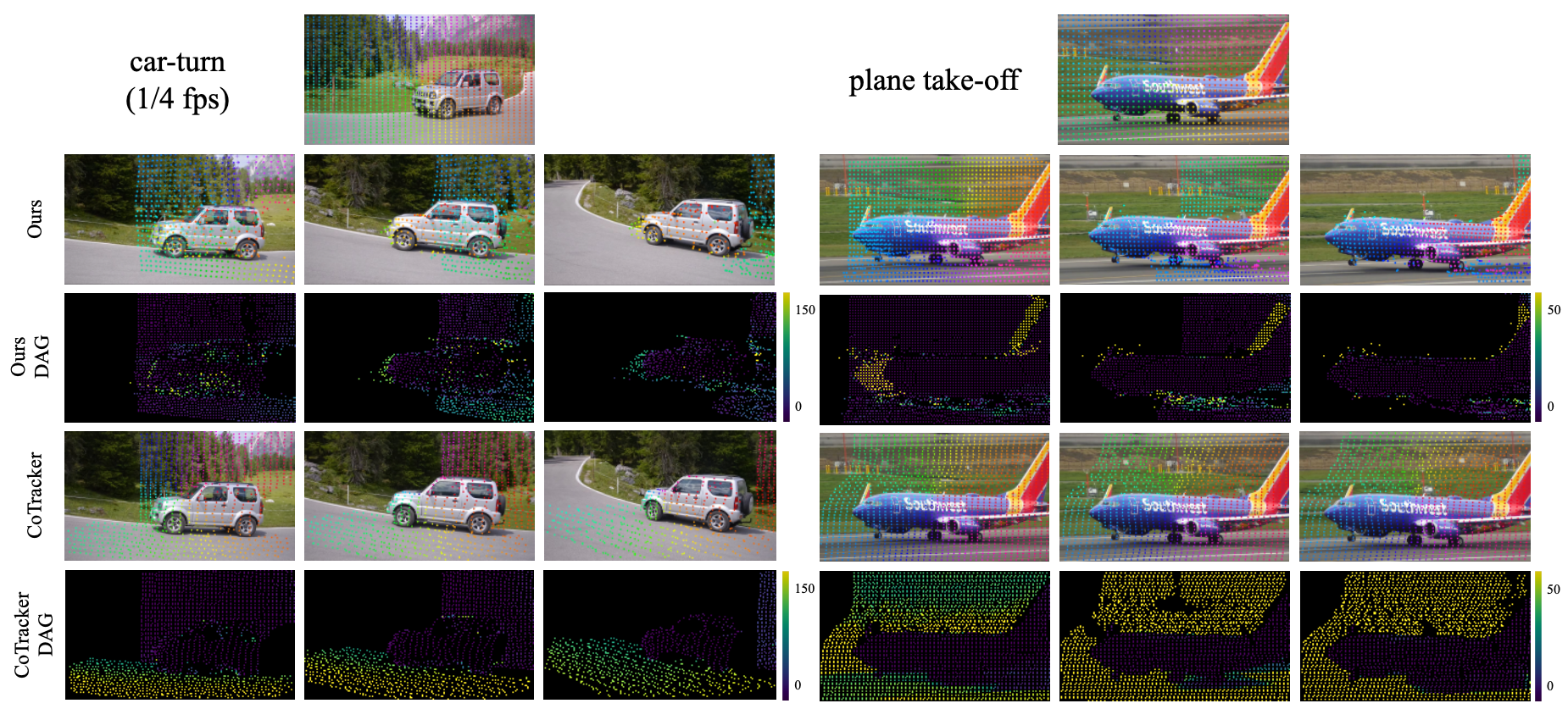}
    \caption{Failure case of CoTracker and visualization of DAG. We track both the foreground and the background pixels. The error map shows the error magnitude of all trajectory points, where bright yellow equals a large error and dark purple equals a small error.}
    \label{fig:cotracker_failure}
    \vspace{-2.5em}
\end{figure}

\begin{table}[t]
\caption{
Comparison of convergence robustness of OmniMotion~\cite{wang2023tracking} and ours.
}
\centering

{{
\small 
\begin{tabular}{p{2.5cm}lccccccccc}
\toprule

\textbf{Method} &  \multicolumn{9}{c}{\textbf{$\delta^x_{avg}$ $\uparrow$}} \\
            & \multicolumn{4}{c}{motocross-jump} & & \multicolumn{4}{c}{libby} \\
            \cmidrule(){2-5} \cmidrule(){7-10}
            &    min & max & mean & std &  &  min & max &  mean & std\\ 
\midrule
Omnimotion  & 4.7 &  60.5 & 26.3 & 26.1         & & 2.3  & 18.0 & 8.86 & 5.9  \\ 
Ours w/o depth  &   4.4 &  65.5 & 44.3 & 23.5   & &  1.8    & 20.2  &   12.7    & 6.6  \\
Ours          & 75.2 & 76.4 &  75.6   &  0.5    & & 40.1 &   48.5  & 45.7   &   3.0 \\
\bottomrule
\end{tabular}
}}%

\label{tab: robust}
\end{table}

\subsubsection{CoTracker~\cite{karaev2023cotracker}} As a learned method, Cotraker works well when the expectation of the learned distribution aligns with that of the testing distribution. Nevertheless, this is not always the case when evaluating videos that have not been previously seen. In Fig.~\ref{fig:cotracker_failure}, we show several cases where Cotracker fails. In Fig.~\ref{fig:cotracker_failure}-Left, when the frame rate is low and significant relative motion exists, we observe that the pixels representing the ground are inaccurately tracked as moving along with the vehicle. 
In Fig.~\ref{fig:cotracker_failure}-Right, we demonstrate that the background points, despite having rich textures, are still inaccurately tracked. The background failure instances are not adequately represented in the DAVIS benchmark because the ground-truth points are labeled as foreground majorly.
In these two sequences, we observe that the local optical flow is significantly more accurate than long-term tracks. Therefore, we further quantitatively measure these failures by computing the average disagreement with the trajectory and optical flow by 
\begin{equation}
    DAG = \frac{1}{|P|}\sum_{( {p}_i,  {p}_j) \in P} ||( {p}_j -  {p}_i) - f_{i\rightarrow j}( {p}_i))||_2,\; j - i = 1
\end{equation}
where $P$ is the set of all visible trajectory points, $( {p}_i,  {p}_j)$ is the two adjacent points on a trajectory and $f_{i\rightarrow j}$ is the flow computed between frame $i, j$. We report these accuracies in Tab~\ref{tab: agreement}. In this case, the less the disagreement, the more precise the track is. We observe that ours still tracks reasonably well. In contrast, Cotraker is not able to predict accurate point tracks in both cases.

\vspace{-1em}
\subsubsection{OmniMotion~\cite{wang2023tracking}}
\begin{figure}[tb]
    \centering
    \includegraphics[width=1\linewidth]{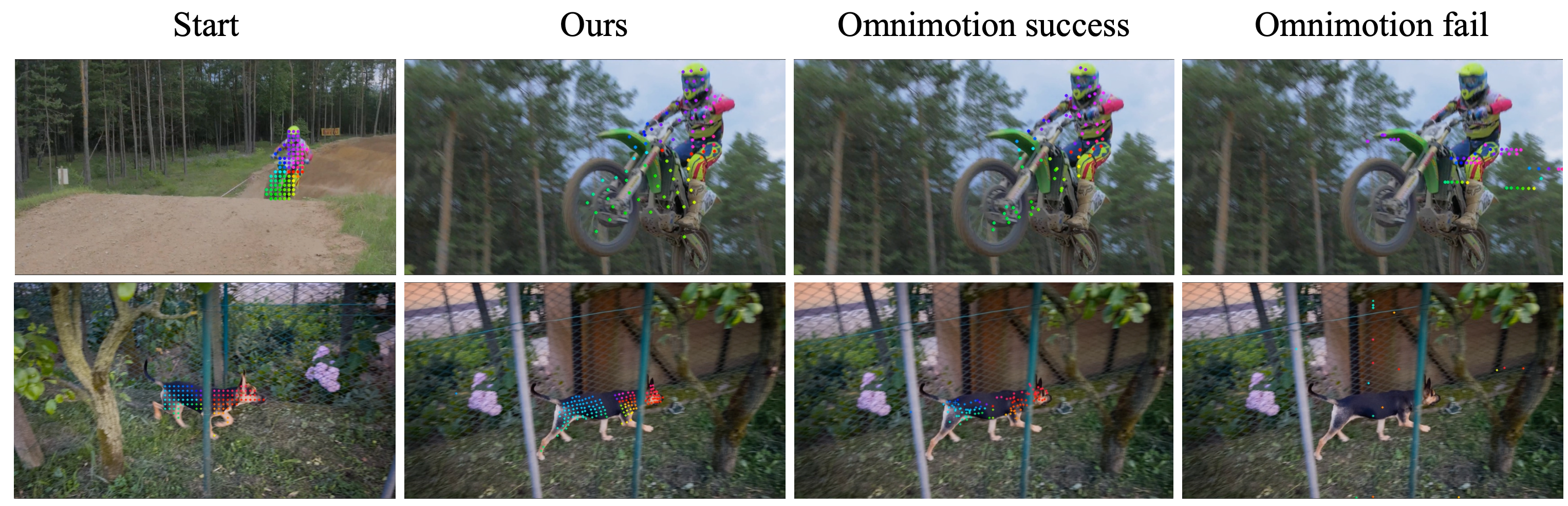}
    \caption{\textbf{Robustness}: running OmniMotion~\cite{wang2023tracking} with different random seeds will result in highly variant fitting results shown on the right while ours is stable.}
    \label{fig: omni_fail}
    \vspace{-1em}
\end{figure}

We further verify one of our important arguments of robustness. When the network is optimized on the same scene with different random seeds, OmniMotion~\cite{wang2023tracking} often results in fitting errors with high variance, as shown in Tab.~\ref{tab: robust} and Fig.~\ref{fig: omni_fail}.
In contrast, our approach demonstrates stability even with varying random seeds, as illustrated in Table~\ref{tab: robust}. Our robustness is primarily attributed to the incorporation of the Depth prior (Sec.~\ref{sec:method_depth}), which acts as a regularization technique and restricts the optimization space. This decision is supported by the findings in Tab.~\ref{tab: robust} when we ablate the impact of removing the depth from our model.

\vspace{-1em}
\section{Conclusion}
\vspace{-1em}
We present a novel approach to computing the long-term trajectories of pixels from a video. Our approach aims to maximize computational efficiency and robustness, which are key shortcomings of the previous work. By proposing a novel invertible block with local grid, we boost the expressivity of the mapping functions. Additionally, we take advantage of recent foundational models and bootstrap long-term semantic consistency with short-term flow consistency. Our model achieves state-of-the-art performance in optimizing the tracking test time, while significantly reducing computational time by $90\%$. Compared with Omnimotion, the previous SoTA, our method tracks everything everywhere faster and more robsustly.

\subsubsection{Acknowledgement} We gratefully acknowledge the financial support through the NSF IIS-RI 2212433 grant, and a gift from AWS AI to Penn Engineering's ASSET Center for Trustworthy AI.

\clearpage  %

\bibliographystyle{splncs04}
\bibliography{main}

\appendix

\large \textbf{Appendix}

\section{CaDeX++ }
\label{sec:cadexpp}

We implemented the temporal feature grid $\Psi_l(i)$ in three resolutions: $T/20$, $T/4$, and $13T/20$, where $T$ is the number of frames. Each resolution has a feature dimension of 16. For the spatial feature grid $\Phi_l(x, y)$, we implemented 2 resolutions 12 and 96, with feature dimensions 32 for each resolution. 2 hidden layers are set for the tiny MLP. We perform ablation studies on DAVIS~\cite{davis} scenes: breakdance, bmx-trees, libby, parkour, and blackswan.

The tiny MLP predicts the positive incremental bias of the control points as $[(\Delta \alpha^1, \Delta \beta ^1)...(\alpha^B, \Delta \beta ^B)]$ together with the positive outlier slope $k_l, k_r$. We divide the incremental bias into two sets $\{(\Delta\alpha_N^i, \Delta\beta_N^i)\}_{i=1}^{B/2}$ and $\{(\Delta\alpha_P^i, \Delta\beta_P^i)\}_{i=1}^{B/2}$ to generate the control points with negative and positive $\alpha$ values. For the control points with  negative $\alpha$ values, their coordinates are computed as:
\begin{equation}
    (\alpha_N^k, \beta_N^k) = - (\sum_{i=1}^k \Delta\alpha_N^i, \sum_{i=1}^k \Delta\beta_N^i)
\end{equation}
While the control points with positive $\alpha$ values are aggregated as:
\begin{equation}
    (\alpha_P^k, \beta_P^k) = (\sum_{i=1}^k \Delta \alpha_P^i, \sum_{i=1}^k \Delta \beta_P^i)
\end{equation}
For the input that lies outside the left-most or right-most control point $(\alpha_m, \beta_m)$, we compute the output as:
\begin{equation}
    z' = k_m(z - \alpha_m) + \beta_m
\end{equation}
where $k_m$ is the outlier slope.

\section{Preparing Long-term Correspondence}
During training, we sample flow for each query frame among a neighbourhood of 12 frames, and search long-term correspondence outside a neighborhood of 10 frames. Coarse correspondences are computed on the low-resolution feature maps of DINOv2~\cite{oquab2023dinov2}. We applied three strong filters to remove noisy and keep representative matches. 
\begin{itemize}
    \item \textbf{Mutual Maximum}. For a matched pair $(p_i, p_j)$ of two frames $F_i, F_j$, the best matching of $p_i$ in frame $F_j$ should be $p_j$ and vice versa: 
    \begin{equation}
        \mathop{\text{argmax}} \limits_{p_i\in F_i}{S \langle \mathop{\text{argmax}} \limits_{p_j\in F_j}{S\langle p_k, p_j \rangle}, p_i  \rangle} = p_k,\; p_k \in F_i
    \end{equation}
    where $S\langle p_i, p_j \rangle$ denotes the cosine similarity between the feature of points $p_i, p_j$. We only choose the pairs that have similarity over $\theta_m = 0.75$.

    \item \textbf{Background Filter}. For a point $p_k$ in a matched pair, we compute the similarity between $p_k$ with all other points in its feature map. Then we count the number of similar points beyond a threshold of $\theta _s$. We keep the points that have less then $N_s$ similar points. We set $\theta_s = 0.55$ and $N_s = 100$.
    \begin{equation}
        \sum_{p_i\in F_i} \mathbf{1}(S \langle p_k, p_k \rangle > \theta_s) < N_s
    \end{equation}
    
    \item \textbf{Local Noise Filter}. For a point $p_k$ in a matched pair, we compute the similarity among its $11 \times 11 $ neighbor points $M(p_k)$ and sum up all the similarity. We choose the points with total local similarity larger than $\theta_l$ = 30.

    \begin{equation}
        \sum_{p_i \in M(p_k)} S \langle p_i, p_k \rangle > \theta_l
    \end{equation}
    
\end{itemize}

\section{Optimization Based on CoTraker}

We utilizes the output of CoTracker as part of our training supervision for each scene. The optimization result on DAVIS dataset is shown in Tab. \ref{tab: cotracker}.

\vspace{10pt}

\begin{table}[]
\caption{
Result of optimization on DAVIS with CoTracker output.
}
\centering

{{
\small 
\begin{tabular}{p{2.5cm}lcccc}
\toprule

\textbf{Method} & \multicolumn{4}{c}{DAVIS~\cite{davis}} \\
            \cmidrule(){2-5}
            &   AJ$\uparrow$ & $\delta^x_{avg}$ $\uparrow$ & OA$\uparrow$  & TC $\downarrow$  \\
\midrule
CoTracker~\cite{karaev2023cotracker}  &   \textbf{65.1}    &   79.0    &   \textbf{89.4}    &   0.93     \\
Ours       & 62.2   & \textbf{80.0} &  86.8   &  \textbf{0.69}  \\
\bottomrule
\end{tabular}
}}%

\label{tab: cotracker}
\end{table}

\section{Limitation and Future Research}

Like other optimization-based methods, the efficacy of our tracking performance is dominated by the precision and quality of the input depth and the pixel correspondence.

Moreover, current network architecture primarily addresses 2D pixel tracking task. It is imperative to investigate its potential capabilities in other tasks, including 3D reconstruction, object pose estimation, and content generation.

\end{document}